\documentclass[10pt,twocolumn,letterpaper]{article}

\usepackage{wacv}
\usepackage{times}
\usepackage{epsfig}
\usepackage{graphicx}
\usepackage{amsmath}
\usepackage{amssymb}
\usepackage{textcomp}
\usepackage{xcolor}
\usepackage{mathrsfs}
\usepackage{booktabs} 
\usepackage{caption}
\usepackage{subcaption} 
\usepackage{booktabs}
\usepackage{multirow}
\usepackage[accsupp]{axessibility} 

\makeatletter
\@namedef{ver@everyshi.sty}{}
\makeatother
\usepackage{dblfloatfix}
\usepackage{bigstrut}
\usepackage{multirow}
\usepackage{color}
\usepackage{pgfplots}
\usepackage{balance}
\usepackage{enumitem}

\usepackage{breqn}
\usepackage{float}

\def\wacvPaperID{****} 

\wacvfinalcopy 

\ifwacvfinal
\def\assignedStartPage{9876} 
\fi

\ifwacvfinal
\usepackage[breaklinks=true,bookmarks=false]{hyperref}
\else
\usepackage[pagebackref=true,breaklinks=true,colorlinks,bookmarks=false]{hyperref}
\fi

\ifwacvfinal
\else
\fi

\begin{document}

\title{Algorithmic Fairness in Face Morphing Attack Detection}
\author{Raghavendra Ramachandra \quad Kiran Raja \quad Christoph Busch\\
Norwegian University of Science and Technology (NTNU), Norway\\ 
	\{\tt\small raghavendra.ramachandra; kiran.raja; christoph busch\} @ntnu.no\\
	
	}

\maketitle
\thispagestyle{empty}
\pagenumbering{gobble}
\begin{abstract}
Face morphing attacks can compromise Face Recognition System (FRS) by exploiting their vulnerability. Face Morphing Attack Detection (MAD) techniques have been developed in recent past to deter such attacks and mitigate risks from morphing attacks. MAD algorithms, as any other algorithms should treat the images of subjects from different ethnic origins in an equal manner and provide non-discriminatory results. While the promising MAD algorithms are tested for robustness, there is no study comprehensively bench-marking their behaviour against various ethnicities. In this paper, we study and present a comprehensive analysis of algorithmic fairness of the existing Single image-based Morph Attack Detection (S-MAD) algorithms. We attempt to better understand the influence of ethnic bias on MAD algorithms and to this extent, we study the performance of MAD algorithms on a newly created dataset consisting of four different ethnic groups. With Extensive experiments using six different S-MAD techniques, we first present benchmark of detection performance and then measure the quantitative value of the algorithmic fairness for each of them using Fairness Discrepancy Rate (FDR). The results indicate the lack of fairness on all six different  S-MAD methods when trained and tested on different ethnic groups suggesting the need for reliable MAD approaches to mitigate the algorithmic bias.    
\end{abstract}

\section{Introduction}
\label{sec:intro}

Face recognition systems are widely deployed in various security-based applications, including border control. While FRS can provide a user-friendly, seamless, and reliable verification solution, they are also known to be vulnerable to different types of attacks, including the recently reported face morphing attacks \cite{Sushma_MAD_Survey,  MargicPass, makrushin2018overview}. Face morphing attacks against FRS are carried out by combining face images of multiple subjects to generate a single image that can be verified equally against all contributing subjects \cite{Sushma_MAD_Survey}.  Existing works on face morphing attacks have reported the vulnerability of FRS towards such attacks  \cite{Raghavendra2016, Scher2017, IWBF2017_StirTrace, raghavendra2017transferable, raghavendra2017face, neubert2017Imagedegradation, seibold2018reflection, RagCVIP2018, RagISBA2019, Ferrara_PrintScan2019, scherhag2020deep, Venkatesh_2020_WACV, DeepResidualNoise_IPTA_2019, Jag_ABC_gate_2019, NISTReport2020, MorphStyleGAN2020, Raja2020MorphingAD, MIPGAN, soleymani2020differential, soleymani2021mutual, damer2021pw, damer2021pw, aghdaie2021attention, banerjee2021conditional}. Extensive benchmark presented by NIST FRVT MORPH \cite{NISTReport2020} and the Bologna Online Evaluation Platform (BOEP) \cite{Raja2020MorphingAD} also indicate the high vulnerability of FRS with respect to face morphing attacks. In addition to FRS, also human observer analysis (including forensic experts, border guards and biometric researchers) has also reported the challenges to detect face morphing attacks \cite{MAD_Human2019, RobertSon2018, MakrushinHuman, Morph_PLUSONE}. These factors have motivated the development of multiple automatic face Morphing Attack Detection (MAD) to detect the evolving morphing attacks, especially tailored for border control applications. 

\begin{figure}[t]
	\begin{center}
		\includegraphics[width=1\linewidth]{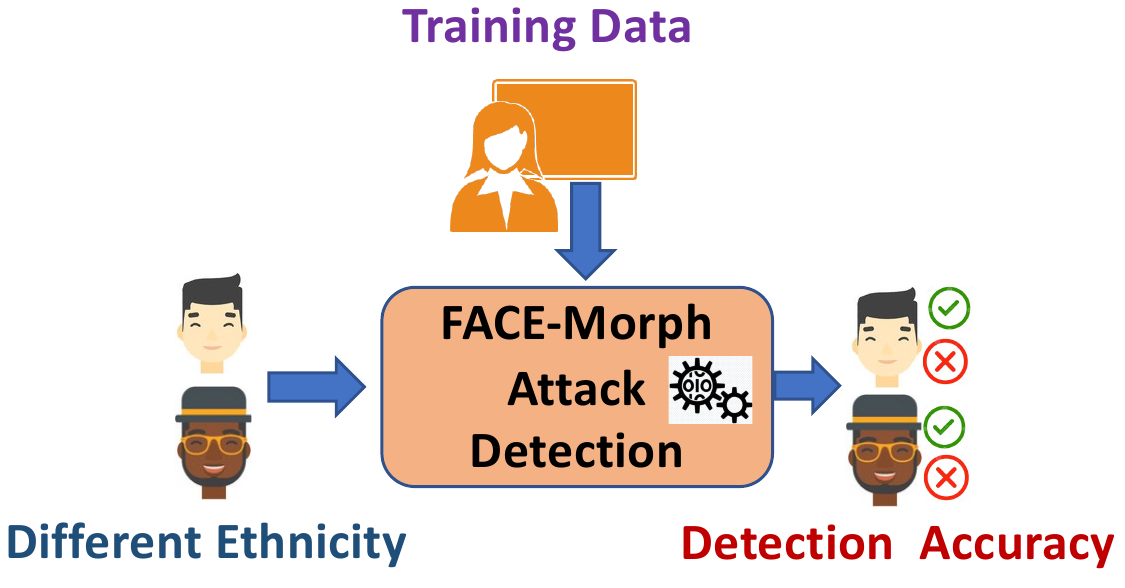}
	\end{center}
	\caption{Morph Attack Detection under the influence of ethnic factors}
	\label{fig:Intro}
\end{figure}

A number of MAD techniques have been proposed in the past five years \cite{damer2021pw}  \cite{aghdaie2021attention}  \cite{RagISBA2019}  \cite{soleymani2020differential}  \cite{Sushma_MAD_Survey} \cite{Venkatesh_2020_WACV} \cite{raghavendra2017transferable} which can be broadly classified as Single image-based MAD (S-MAD) and Differential based MAD (D-MAD) \cite{Sushma_MAD_Survey}. The S-MAD techniques use a single image on which morphing detection is performed and is deemed suitable for forensic applications. The D-MAD technique is based on two images to detect the morphing attacks and is more suitable for Automatic Border Control (ABC) gates used in border control scenario. Irrespective of the MAD type, most of the existing MAD methods are based on machine learning, where decisions are made based on the trained data to obtain a MAD algorithm. 

\begin{figure*}[b]
	\begin{center}
		\includegraphics[width=0.9\linewidth]{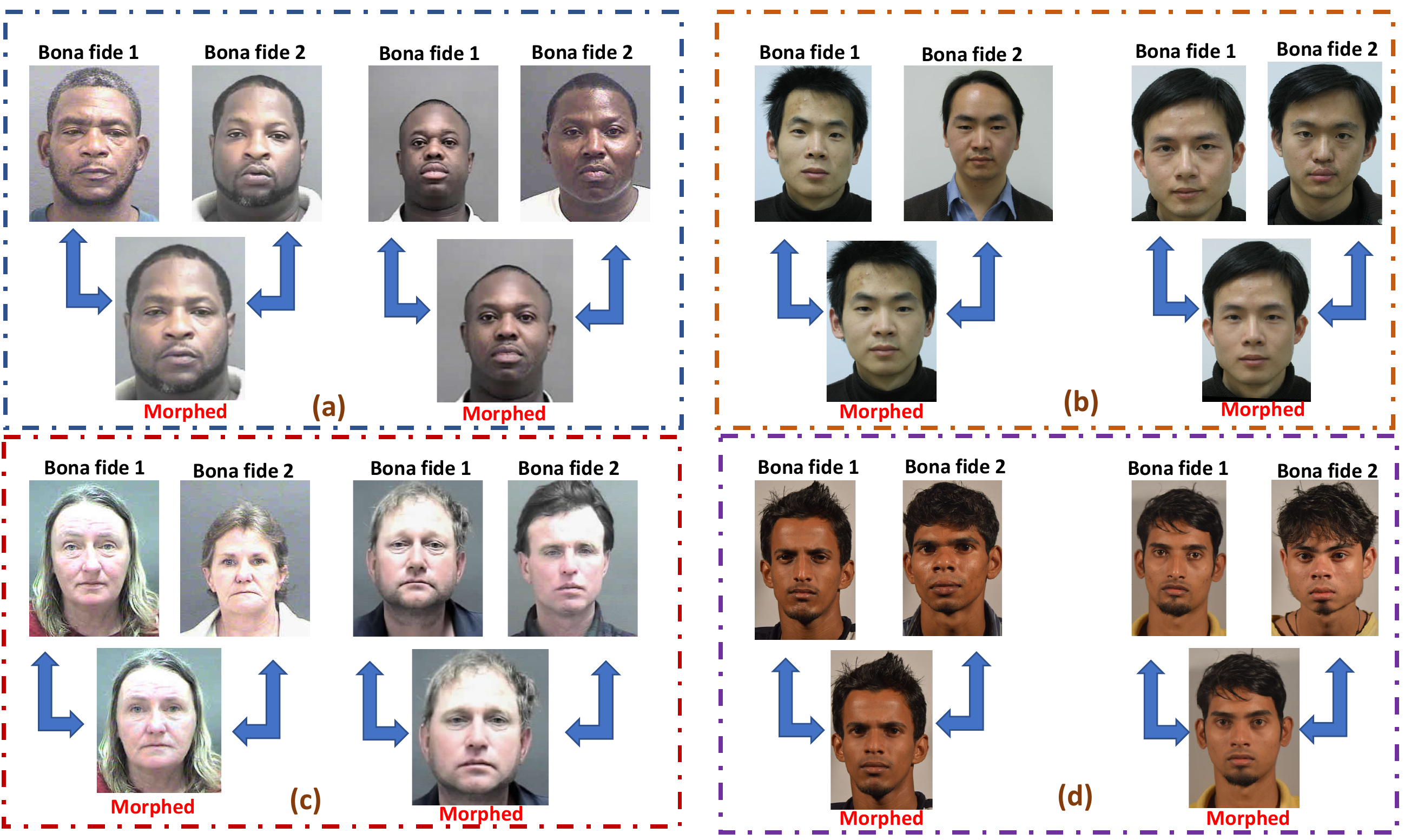}
	\end{center}
	\caption{Illustration of samples from MEM dataset: (a) Ethnicity-1: African (b) Ethnicity-2: Chines (c) Ethnicity-3: Caucasian (d) Ethnicity-4: Indian}
	\label{fig:DB}
\end{figure*}

While it is well established that learning-based systems are sensitive to various factors including ageing, ethnicity, gender, data capture environment making them involuntarily biased in making decisions \cite{mehrabi2021survey}, similar works in FRS have also established the same shortcoming of bias in FRS \cite{howard2020quantifying}, \cite{robinson2020face}, \cite{terhorst2021comprehensive}. Such bias factors prevent in achieving demographic fairness to groups such as ethnic origin or gender causing social, legal and ethical implications. 

Taking the analogy from demographic studies on other functionality of FRS such as face segmentation or recognition as such, one can assume the potential biases in MAD techniques which employ face images exclusively in learning algorithms. As most MAD methods are largely learning based, they are also naturally sensitive to various factors like ageing, gender, and ethnicity bias. For instance,  the influence of ageing on MAD was studied in Venkatesh et al. \cite{Venkatesh_2020_IJCB}, the influence of gender on MAD techniques was studied in Raghavendra et al. \cite{RagISBA2019}.  Earlier reported results indicate a lower influence of ageing on MAD performance where two different age groups with an age difference of up to two years and five years were analyzed using different  MAD techniques \cite{Venkatesh_2020_IJCB}. However, gender was shown to have some influence on MAD,  especially on the texture-based MAD methods \cite{RagISBA2019} where MAD approaches detected male morphed images relatively better as compared to female morphed images.  The challenges to detect the female morphing images was also demonstrated in the extensive evaluation reported by Raja et al. \cite{Raja2020MorphingAD} throwing light on role of gender. 


Although these studies, to a certain extent, indicate the influence of various factors on the detection performance of the MAD techniques, the  focus of investigation was limited to only one aspect. To the best of our knowledge, there exists no systematic work on benchmarking any algorithmic bias with respect to distribution of ethnicity of data (both training and testing) in S-MAD techniques.  The influence of ethnicity on MAD techniques was briefly discussed in \cite{Raja2020MorphingAD}, that indicates the sensitivity of the  MAD performance to different ethnicity biases. However, there is no detailed analysis on benchmarking the performance of MAD techniques on different ethnicity groups, mainly due to limited distribution of ethnicity in the dataset used for benchmark. Noting these limitations, we aim to seek answers to the following research questions in an attempt to benchmark the algorithmic fairness of MAD algorithms: 
\begin{itemize}[leftmargin=*,noitemsep, topsep=0pt,parsep=0pt,partopsep=0pt]
  \item  \textbf{RQ\#1:} Does the ethnicity influence the detection performance of S-MAD algorithms?
    \item  \textbf{RQ\#2:}  For which ethnic groups is the detection most challenging? 
     \item  \textbf{RQ\#3:}  What category of S-MAD algorithms (deep learning/hand-crafted features/morphing noise based) degrade in performance as a result of ethnic bias?
\end{itemize}

We answer the above mentioned questions to establish the impact of ethnicity on MAD by training and testing a set of six MAD algorithms using Mixed Ethnicity Morphed (MEM) datasets consisting of morphed images from four different ethnic groups. We limit the scope of this work on S-MAD algorithms alone by considering its wide application in the passport application process, where (printed) face images are submitted, and the limitations of S-MAD to detect the the morphing attacks \cite{Raja2020MorphingAD}.

In this work, we address the posed research questions in a systematic manner through the following contributions:
\begin{itemize}[leftmargin=*,noitemsep, topsep=0pt,parsep=0pt,partopsep=0pt]
    \item The first key contribution is the creation of the Mixed Ethnicity Morph (MEM) dataset consisting of bona fide images and corresponding morphed images from four different ethnic groups. The four diverse ethnic group includes (1) African ethnicity (2) Caucasian ethnicity (3) Chinese ethnicity and (4) Indian ethnicity. 
    \item The second key contribution is the benchmarking of six different S-MAD techniques with three different evaluation protocols. The first protocol is designed to evaluate the intra-ethnicity performance of MAD algorithms, the second protocol is to study the cross-ethnicity performance variation of MAD algorithms and the third protocol is a combined performance of MAD algorithm across ethnicity. All the three protocols study the algorithmic bias as a result of ethnicity distribution in training data for MAD algorithm. 
    \item The third contribution is the benchmarking of quantitative fairness of MAD techniques using Fairness Discrepancy Rate (FDR) metric. 
\end{itemize}
In the rest of this paper, Section \ref{sec:DB} presents the dataset preparation and the morphing process used to construct  four different ethnic datasets. Section \ref{sec:MAD} discuss the S-MAD techniques employed in this work. Section \ref{sec:exp} discuss the experiments and results and Section \ref{sec:conc} draws the conclusion. 
\section{Mixed Ethnicity Morph (MEM) Dataset}
\label{sec:DB}
Noting the non availability of morphed datasets with different ethnicity distribution, we introduce the newly created Mixed Ethnicity Morph (MEM) dataset in this section. We specifically focus on four ethnic origins grouped as (a) Ethnicity-1: African (b) Ethnicity-2: Chinese (c) Ethnicity-3: Caucasian (d) Ethnicity-4: Indian. The choice of these four-ethnicity groups is based on the diverse variation in the population that can provide insights on performance of the S-MAD techniques. In the following section, we briefly describe the different subgroups of the MEM dataset.  
\paragraph{Ethnicity-1 dataset:} This dataset represents the African ethnicity and is derived from the MORPH II non-commercial dataset \cite{MORPHII_DB}  that is publicly available. MORPH II dataset consists of male and female data subjects from different ethnicity that includes African, European, Asian and Hispanic. We select the African ethnicity data subjects from the MORPH II dataset. Ethnicity-1 set consists of 540 data subjects representing the African population in MEM dataset. 
\vspace{-0.5cm}
\paragraph{Ethnicity-2 dataset:} This dataset represents the Chinese ethnicity and is derived from CASIA VIS-NIR-3D dataset \cite{ CASIAFAceDB}. This dataset has 100 unique data subjects captured in the enrollment quality (frontal pose, uniform lighting and etc.).  
\vspace{-0.5cm}
\paragraph{Ethnicity-3 dataset:} This dataset represents the Caucasian ethnicity and is derived from the MORPH II non-commercial dataset \cite{MORPHII_DB}. We have selected both European and Hispanic ethnicity that corresponds to 100 unique data subjects. 
\vspace{-0.5cm}
\paragraph{Ethnicity-4 dataset:} This dataset represents the Indian ethnicity, especially from the southern India. This is a newly collected dataset in the controlled conditions by simulating the passport enrolment conditions with uniform lighting, zero expressions and only frontal pose. This dataset is comprised of 150 data subjects. 

The MEM dataset consists of $24874$ images including bona fide and morphed images, and the distribution of the total images is provided in Table \ref{tab:statsTable}.

\begin{table}[htbp]
  \centering
  \caption{Statistics of MEM Dataset}
  	\resizebox{01\linewidth}{!}{
    \begin{tabular}{lccccc}
    \hline
    Database & \multicolumn{2}{c}{No. of training images} & \multicolumn{2}{c}{No. of Testing images} & \multicolumn{1}{c}{\multirow{2}[4]{*}{Total}} \bigstrut\\
\cline{2-5}     & Bona fide & Morphed & Bona fide & Morphed &  \bigstrut\\
    \hline
    Ethnicity-1 & 607 & 607 & 881 & 18797 & 20892 \bigstrut\\
    Ethnicity-2 & 151 & 151 & 147 & 1176  & 1625 \bigstrut\\
    Ethnicity-3 & 102 & 102 & 145 & 948   & 1297 \bigstrut\\
   Ethnicity-4 & 119 & 119 & 151 & 671   & 1060 \bigstrut\\
    \hline
    \multicolumn{5}{c}{}  Total Images              & 24874 \bigstrut\\
    \hline
    \end{tabular}%
    }
  \label{tab:statsTable}%
\end{table}%

\begin{table*}[htbp]
	\centering
	\caption{Quantitative results of MAD algorithms when trained on Ethnicity-1 and tested on the rest}
	\resizebox{1\linewidth}{!}{
\begin{tabular}{lcccccccccccc}
	\hline
	& \multicolumn{3}{c}{\textbf{Ethnicity-1}} & \multicolumn{3}{c}{\textbf{Ethnicity-2}} & \multicolumn{3}{c}{\textbf{Ethnicity-3}} & \multicolumn{3}{c}{\textbf{Ethnicity-4}}\\
	\hline
	 MAD Algorithms & D-EER (\%) & \multicolumn{2}{c}{BPCER @ APCER} & D-EER (\%) & \multicolumn{2}{c}{BPCER @ APCER} & D-EER (\%) & \multicolumn{2}{c}{BPCER @ APCER} & D-EER (\%) & \multicolumn{2}{c}{BPCER @ APCER}\bigstrut\\
	 &       & =10\%  & =5\% &       & =10\%  & =5\% &       & =10\%  & =5\% &       & =10\%  & =5\% \bigstrut\\
	\hline
	\hline

	Hybrid Features  \cite{RagISBA2019} 
	& 7.41  & 5.78  & 11.23
	& 29.16 & 61.26 & 76.87
	& 22.14 & 29.65 & 38.67 
	& 38.43 & 73.5  & 79.47
	\bigstrut\\

	Ensemble Features \cite{EnsembleFeatures_2020} 
	& 9.98  & 9.89  & 14.98
	& \textbf{27.93} & \textbf{47.61} & \textbf{61.9}
	& \textbf{13.96} & \textbf{20} & \textbf{43.44}
	& 37.14 & 76.15 & 84.14
	\bigstrut\\
	
	Colour denoise \cite{DeepResidualNoise_IPTA_2019} 
	& 12.93 & 16.68 & 27.8
	& 38.75 & 72.14 & 82.99
	& 20.14 & 33.1  & 48.27
	& 47.68 & 84.76 & 91.39
	\bigstrut\\
	
	Residual Noise \cite{Venkatesh_2020_WACV} 
	& \textbf{6.59} & \textbf{5.14} & \textbf{9.17}
	& 33.33 & 63.26 & 60.15
	&  27.45 & 55.86 & 73.16
	& \textbf{26.43} & \textbf{52.98} & \textbf{43.7}
	\bigstrut\\
	
	AlexNet \cite{Venkatesh_2020_WACV} 
	& 7.74  & 5.67  & 9.87
	& 29.25 & 56.46 & 64.62
	& 20.52 & 34.48 & 49.65
	& 35.76 & 66.88 & 72.84
	\bigstrut\\
	
	ResNet101 \cite{Venkatesh_2020_WACV} 
	& 8.96	&	8.05	& 13.05
	& 31.97 & 61.22 & 75.51 
	& 14.46 & 17.24 & 26.89 
	& 29.61 & 63.57 & 77.48
	\bigstrut\\
	\hline			
\end{tabular}%
}
\label{tab:TableEh1}%
\end{table*}%

\begin{table*}[htbp]
	\centering
  \caption{Quantitative results of MAD algorithms when trained on Ethnicity-2 and tested on the rest}
	\resizebox{1\linewidth}{!}{
\begin{tabular}{lcccccccccccc}
	\hline
	& \multicolumn{3}{c}{\textbf{Ethnicity-1}} & \multicolumn{3}{c}{\textbf{Ethnicity-2}} & \multicolumn{3}{c}{\textbf{Ethnicity-3}} & \multicolumn{3}{c}{\textbf{Ethnicity-4}}\\
	\hline
	 MAD Algorithms & D-EER (\%) & \multicolumn{2}{c}{BPCER @ APCER} & D-EER (\%) & \multicolumn{2}{c}{BPCER @ APCER} & D-EER (\%) & \multicolumn{2}{c}{BPCER @ APCER} & D-EER (\%) & \multicolumn{2}{c}{BPCER @ APCER}\bigstrut\\
	 &       & =10\%  & =5\% &       & =10\%  & =5\% &       & =10\%  & =5\% &       & =10\%  & =5\% \bigstrut\\
	\hline
	\hline

	Hybrid Features  \cite{RagISBA2019} 
	& 24.51 & 42.22 & 54.71
	& 6.2   & 3.4   & 7.48
	&  30.52 & 57.93 & 68.96
	& 31.14 & 53.64 & 53.64
	\bigstrut\\

	Ensemble Features \cite{EnsembleFeatures_2020} 
	& 27.69 & 50.62 & 64.35
	& \textbf{4.13} & \textbf{0.68} & \textbf{2.14}
	& 33.72 & 74.48 & 81.37 
	& 30.98 & 62.25 & 71.52
	\bigstrut\\
	
	Colour denoise \cite{DeepResidualNoise_IPTA_2019} 
	&  25.16 & 45.17 & 60.84
	& 4.93  & 2.14  & 4.76 
	& 34.48 & 73.14 & 87.58 
	& 35.13 & 68.87 & 78.8
	\bigstrut\\
	
	Residual Noise \cite{Venkatesh_2020_WACV} 
	& 29.77 & 52.55 & 63.45
	& 14.76 & 23.12 & 33.33
	& 29.54 & 46.2  & 57.24 
	& \textbf{20.54} & \textbf{31.78} & \textbf{47.15}
	\bigstrut\\
	
	AlexNet \cite{Venkatesh_2020_WACV} 
	& \textbf{22.46} & \textbf{34.61} & \textbf{44.26} 
	& 10.92 & 11.56 & 19.14
	& \textbf{22.18} & \textbf{41.37} & \textbf{49.65}
	&  26.5  & 58.94 & 69.53
	\bigstrut\\
	
	ResNet101 \cite{Venkatesh_2020_WACV} 
	& 34.69 & 65.71 & 74.34
	& 14.96 & 21.18 & 29.25
	& 35.86 & 68.27 & 76.55
	& 31.76 & 60.26 & 69.53 
	\bigstrut\\
	\hline			
\end{tabular}%
}
  \label{tab:TableEh2}%
\end{table*}%

\subsection{Face Morphing Generation}
The face morphing generation can be performed  using either landmark based on image wrapping and blending \cite{UBO_Morphing_Tool, Ferrara2016, Raja2020MorphingAD} or deep learning-based using Generative Adversarial Networks (GAN) \cite{MIPGAN}. In this work, we choose the landmark based face morphing generation from Ferrara et al. \cite{UBO_Morphing_Tool, Ferrara2016, Raja2020MorphingAD} with the morphing factor of $0.5$ as it is well demonstrated to generate a high quality face morphing image that can pose high quality attacks against deep-learning based and Commercial-Off-The-Shelf (COTS) FRS \cite{MIPGAN, Venkatesh_2020_IJCB, damer2021regenmorph}.

The face morphing is carried out between two data subjects by considering its use-case in real-life scenarios of passport application. The morphing pairs are selected following the process mentioned in \cite{RightFaces_Naser_2019, Venkatesh_2020_IJCB} that ensure  to generate high quality face morphing image. Further, the whole dataset is partitioned into two disjoint sets with non-overlapping data subjects such as training and testing sets to benchmark the performance of the MAD techniques. Table \ref{tab:statsTable} indicates the statistics of the MEM dataset, and Figure \ref{fig:DB} shows the example face images corresponding to the different ethnicity from the MEM dataset. 

\begin{table*}[htbp]
	\centering
    \caption{Quantitative results of MAD algorithms when trained on Ethnicity-3 and tested on the rest}
	\resizebox{1\linewidth}{!}{
\begin{tabular}{lcccccccccccc}
	\hline
	& \multicolumn{3}{c}{\textbf{Ethnicity-1}} & \multicolumn{3}{c}{\textbf{Ethnicity-2}} & \multicolumn{3}{c}{\textbf{Ethnicity-3}} & \multicolumn{3}{c}{\textbf{Ethnicity-4}}\\
	\hline
	 MAD Algorithms & D-EER (\%) & \multicolumn{2}{c}{BPCER@APCER} & D-EER (\%) & \multicolumn{2}{c}{BPCER@APCER} & D-EER (\%) & \multicolumn{2}{c}{BPCER@APCER} & D-EER (\%) & \multicolumn{2}{c}{BPCER @ APCER}\bigstrut\\
	 &       & =10\%  & =5\% &       & =10\%  & =5\% &       & =10\%  & =5\% &       & =10\%  & =5\% \bigstrut\\
	\hline
	\hline

	Hybrid Features  \cite{RagISBA2019} 
	& 19.49 & 31.44 & 43.92
	& 23.17 & 41.49 & 51.73 
	& 27.5  & 64.13 & 77.93 
	& 38.91 & 76.15 & 86.16
	\bigstrut\\

	Ensemble Features \cite{EnsembleFeatures_2020} 
	& 18.38 & 28.94 & 40.16 
	& \textbf{21.14} & \textbf{38.77} & \textbf{49.65}
	& 16.55 & 27.58 & 40.68
	& 39.83 & 71.52 & 82.11
	\bigstrut\\
	
	Colour denoise \cite{DeepResidualNoise_IPTA_2019} 
	& 22.46 & 39.38 & 53.91 
	& 25.17 & 46.25 & 58.5 
	& 22.82 & 47.58 & 64.13 
	& 54.97 & 84.15 & 94.16
	\bigstrut\\
	
	Residual Noise \cite{Venkatesh_2020_WACV} 
	& 17.22 & 22.36 & 31.32
	& 32.66 & 55.78 & 65.98
	& 15.84 & 23.44 & 31.72
	& \textbf{23.84} & \textbf{46.35} & \textbf{60.92} 
	\bigstrut\\
	
	AlexNet \cite{Venkatesh_2020_WACV} 
	& \textbf{16.46} & \textbf{22.24} & \textbf{33.48}
	& 33.33 & 56.46 & 61.9
	& \textbf{14.3} & \textbf{21.37} & \textbf{35.86}
	& 30.35 & 56.96 & 68.87
	\bigstrut\\
	
	ResNet101 \cite{Venkatesh_2020_WACV} 
	& 21.34 & 33.71 & 46.18
	& 42.68 & 82.93 & 92.51 
	& 21.23 & 34.48 & 44.13 
	& 49.64 & 80.75 & 86.72 
	\bigstrut\\
	\hline			
\end{tabular}%
}
  \label{tab:TableEh3}%
\end{table*}%

\begin{table*}[htbp]
	\centering
  \caption{Quantitative results of MAD algorithms when trained on Ethnicity-4 and tested on the rest}
	\resizebox{1\linewidth}{!}{
\begin{tabular}{lcccccccccccc}
	\hline
	& \multicolumn{3}{c}{\textbf{Ethnicity-1}} & \multicolumn{3}{c}{\textbf{Ethnicity-2}} & \multicolumn{3}{c}{\textbf{Ethnicity-3}} & \multicolumn{3}{c}{\textbf{Ethnicity-4}}\\
	\hline
	 MAD Algorithms & D-EER (\%) & \multicolumn{2}{c}{BPCER @ APCER} & D-EER (\%) & \multicolumn{2}{c}{BPCER @ APCER} & D-EER (\%) & \multicolumn{2}{c}{BPCER @ APCER} & D-EER (\%) & \multicolumn{2}{c}{BPCER @ APCER}\bigstrut\\
	 &       & =10\%  & =5\% &       & =10\%  & =5\% &       & =10\%  & =5\% &       & =10\%  & =5\% \bigstrut\\
	\hline
	\hline

	Hybrid Features  \cite{RagISBA2019} 
	& 34.65 & 61.97 & 73.55
	& 35.37 & 65.98 & 78.24
	& 33.84 & 66.89 & 76.55
	& 12.62 & 21.19 & 25.82 
	\bigstrut\\

	Ensemble Features \cite{EnsembleFeatures_2020} 
	& 40.43 & 72.19 & 81.27
	& 41.21 & 77.55 & 84.35
	& 41.97 & 79.31 & 84.13
	& 15.92 & 22.51 & 31.78
	\bigstrut\\
	
	Colour denoise \cite{DeepResidualNoise_IPTA_2019} 
	& 39.62 & 72.75 & 80.36
	& 34.14 & 63.16 & 78.23 
	& 46.4  & 84.82 & 91.72
	& 15.21 & 19.2  & 27.15
	\bigstrut\\
	
	Residual Noise \cite{Venkatesh_2020_WACV} 
	& \textbf{27.32} & \textbf{44.38} & \textbf{56.98}
	& \textbf{29.25} & \textbf{44.21} & \textbf{52.38}
	& \textbf{22.14} & \textbf{38.15} & \textbf{51.32}
	& \textbf{11.29} & \textbf{14.56} & \textbf{19.2}
	\bigstrut\\
	
	AlexNet \cite{Venkatesh_2020_WACV} 
	& 26.68 & 46.65 & 57.15
	& 32.17 & 55.18 & 61.22
	& 25.52 & 40.68 & 51.43
	& 13.14 & 19.2  & 29.13 
	\bigstrut\\
	
	ResNet101 \cite{Venkatesh_2020_WACV} 
	& 36.3  & 69.69 & 79.56
	& 48.98 & 85.45 & 91.83
	& 33.77 & 68.96 & 84.13
	& 19.21 & 35.76 & 58.27
	\bigstrut\\
	\hline			
\end{tabular}%
}
  \label{tab:TableEh4}%
\end{table*}%

\begin{table*}[htbp]
	\centering
	\caption{Quantitative results of MAD algorithms when trained and tested on all ethnic datasets}
	\resizebox{1\linewidth}{!}{
		\begin{tabular}{lcccccccccccc}
			\hline
	        & \multicolumn{3}{c}{\textbf{Ethnicity-1}} & \multicolumn{3}{c}{\textbf{Ethnicity-2}} & \multicolumn{3}{c}{\textbf{Ethnicity-3}} & \multicolumn{3}{c}{\textbf{Ethnicity-4}}\\
			\hline
			MAD Algorithms & D-EER (\%) & \multicolumn{2}{c}{BPCER @ APCER} & D-EER (\%) & \multicolumn{2}{c}{BPCER @ APCER} & D-EER (\%) & \multicolumn{2}{c}{BPCER @ APCER} & D-EER (\%) & \multicolumn{2}{c}{BPCER @ APCER}\bigstrut\\
			&       & =10\%  & =5\% &       & =10\%  & =5\% &       & =10\%  & =5\% &       & =10\%  & =5\% \bigstrut\\
			\hline
			\hline
			
			Hybrid Features  \cite{RagISBA2019} 
			& 7.49  & 6.19  & 10.55
			& \textbf{6.48} & \textbf{6.12} & \textbf{7.48} 
			& 16.55 & 30.34 & 44.82
			& 17.88 & 27.15 & 39.76
			\bigstrut\\
			
			Ensemble Features \cite{EnsembleFeatures_2020} 
			&  9.32  & 8.62  & 14.89
			& 7.48  & 6.82  & 8.84
			& 14.46 & 17.93 & 24.13
			& 21.72 & 37.15 & 47.68 
			\bigstrut\\
			
			Colour denoise \cite{DeepResidualNoise_IPTA_2019} 
			& 13.85 & 19.52 & 31.55
			& 14.28 & 16.32 & 34.69
			& 24.75 & 54.48 & 65.51
			& 27.84 & 50.99 & 64.9
			\bigstrut\\
			
			Residual Noise \cite{Venkatesh_2020_WACV} 
			& \textbf{6.95} & \textbf{5.33} & \textbf{9.19}
			& 15.64 & 21.76 & 27.21
			& 11.71 & 13.79 & 16.55 
			& \textbf{13.88} & \textbf{17.88} & \textbf{22.51} 
			\bigstrut\\
			
			AlexNet \cite{Venkatesh_2020_WACV} 
			& 8.94  & 7.83  & 11.69
			& 13.6  & 15.64 & 23.13
			& \textbf{10.49} & \textbf{11.72} & \textbf{21.37}
			& 17.88 & 23.84 & 46.35
			\bigstrut\\
			
			ResNet101 \cite{Venkatesh_2020_WACV} 
			& 9.65  & 9.53  & 15.43
			& 15.56 & 26.53 & 42.85 
			& 15.84 & 23.44 & 35.86
			& 18.36 & 30.46 & 54.96
			\bigstrut\\
			\hline			
		\end{tabular}%
	}
	\label{tab:TableEh5}%
\end{table*}%

\begin{figure*}[htbp]
\begin{subfigure}{.245\textwidth}
  \centering
  \includegraphics[width=1\linewidth]{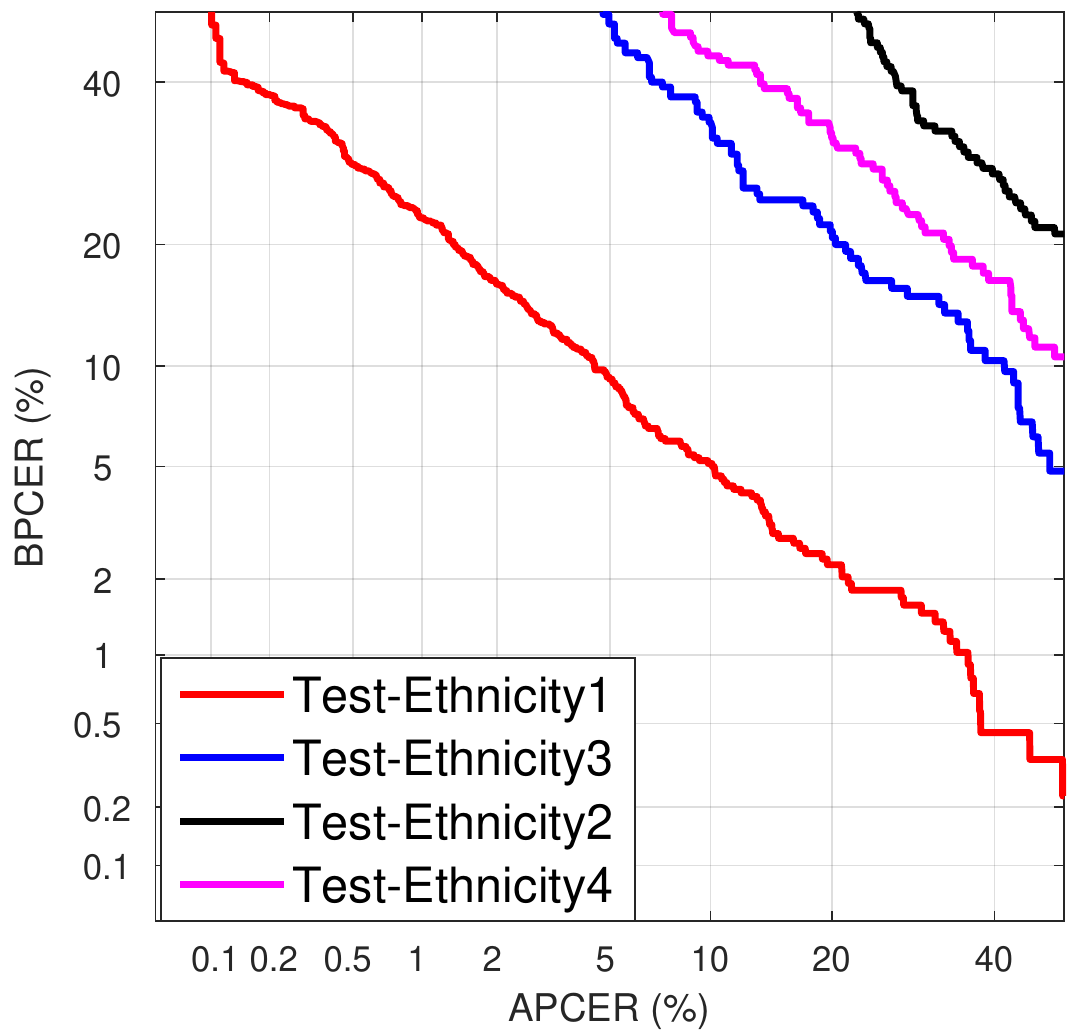}  
  \caption{Train: Ethnicity-1}
  \label{fig:sub-first}
\end{subfigure}
\begin{subfigure}{.245\textwidth}
  \centering
  \includegraphics[width=1\linewidth]{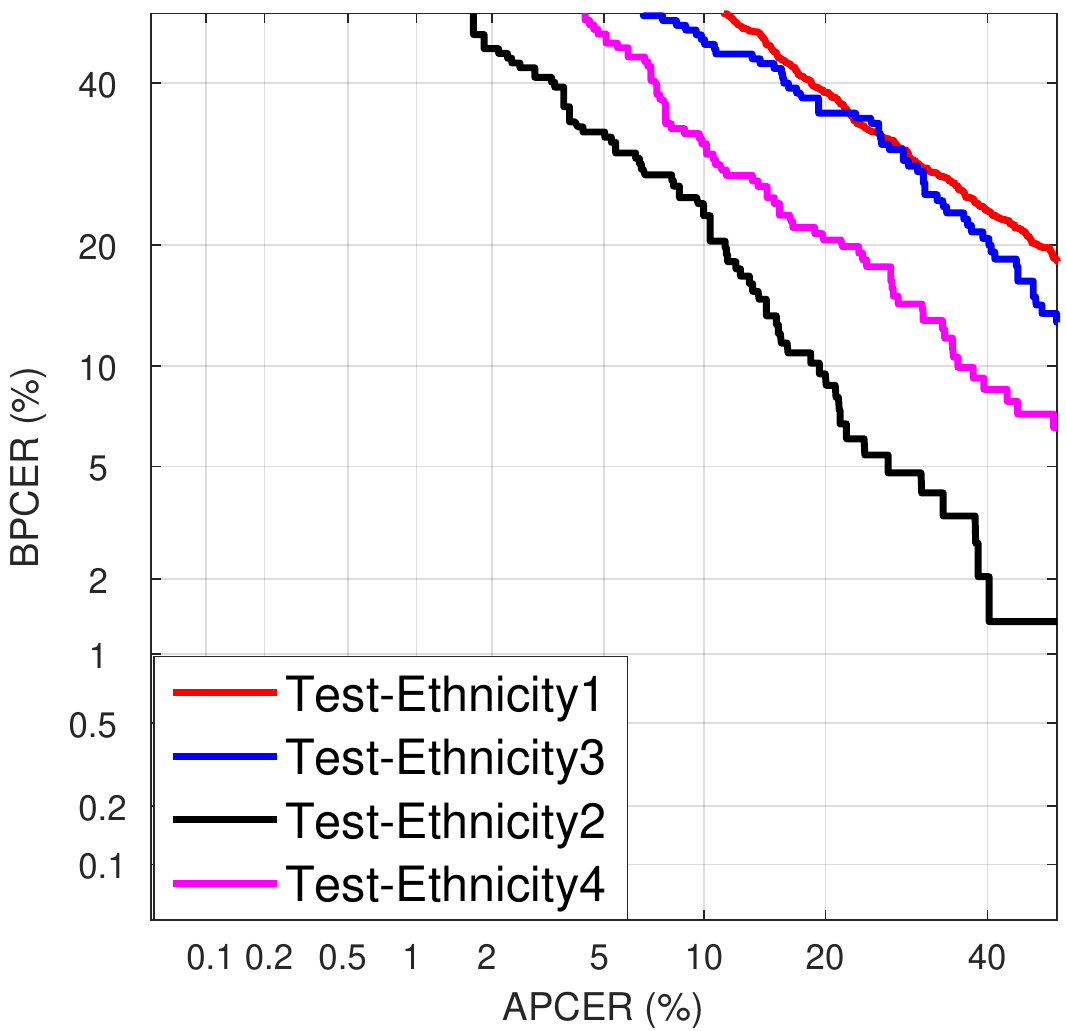}  
  \caption{Train: Ethnicity-2}
  \label{fig:sub-second}
\end{subfigure}
\begin{subfigure}{.245\textwidth}
  \centering
  \includegraphics[width=1\linewidth]{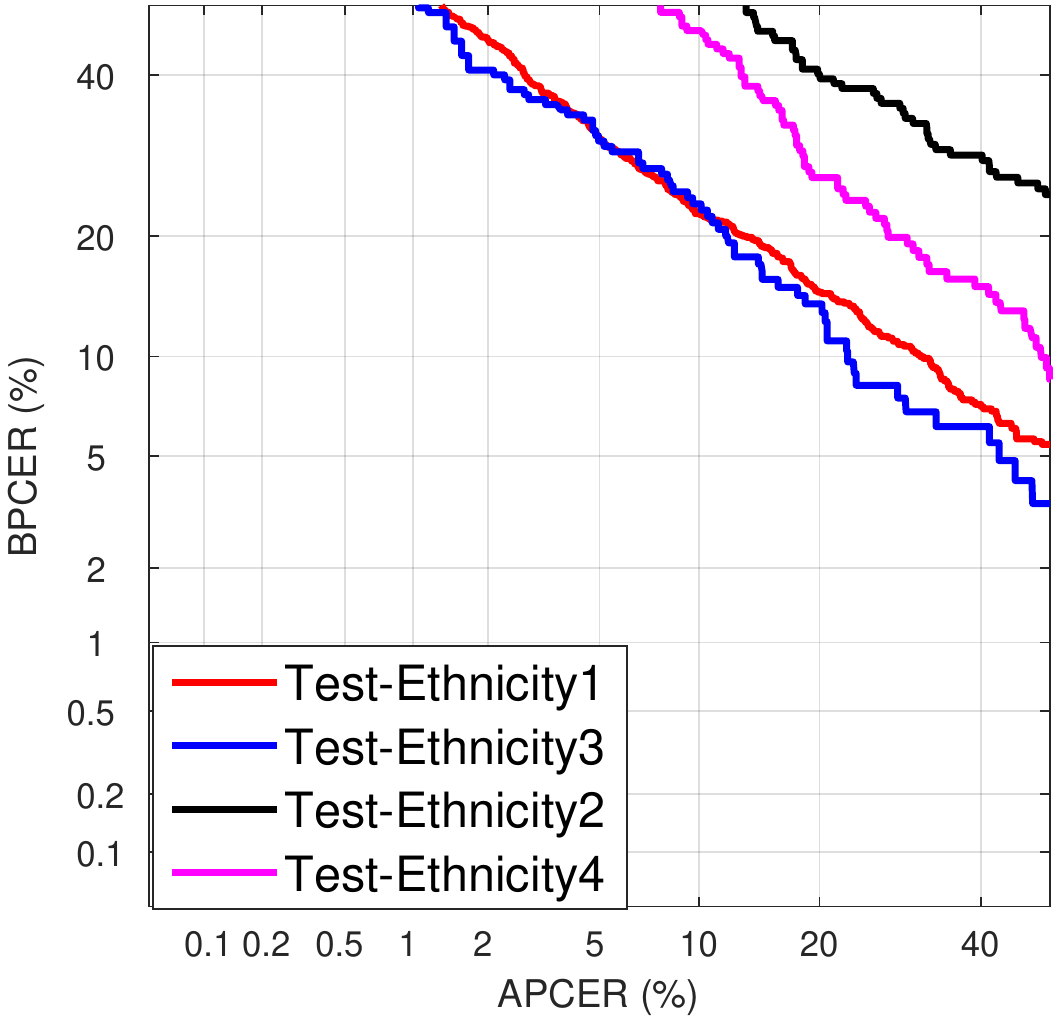}  
  \caption{Train: Ethnicity-3}
  \label{fig:sub-second}
\end{subfigure}
\begin{subfigure}{.245\textwidth}
  \centering
  \includegraphics[width=1\linewidth]{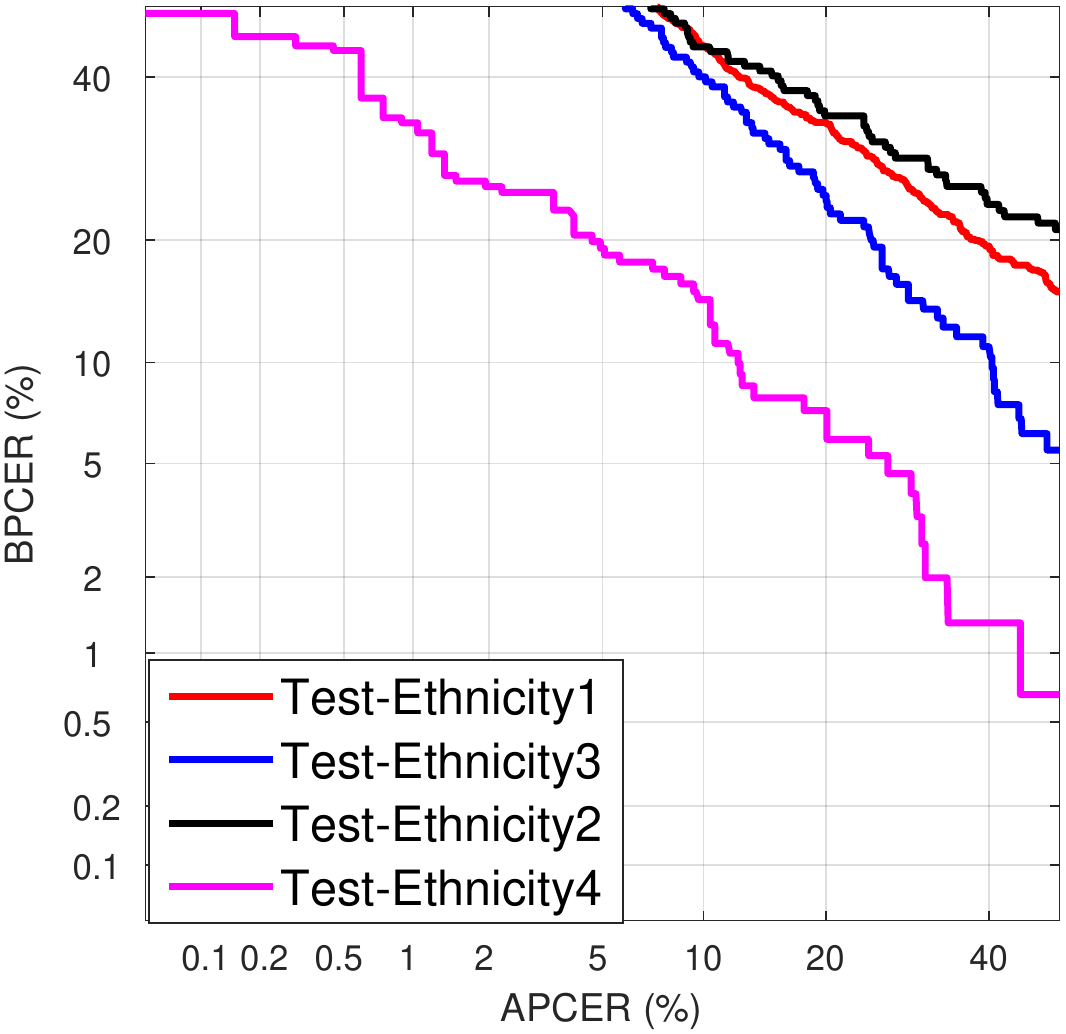}  
  \caption{Train: Ethnicity-4}
  \label{fig:sub-second}
\end{subfigure}
\caption{DET plots indicating the performance of Residual Noise \cite{Venkatesh_2020_WACV}}
\label{fig:Plot1}
\end{figure*}

\section{Face Morph Attack Detection Methods}
\label{sec:MAD}
In order to study the algorithmic fairness, we choose six different state-of-the-art S-MAD techniques that can be broadly classified into (1) deep feature-based (2) hand-crafted features (3) morphing noise based MAD. All these algorithms are selected based on the morph attack detection accuracy reported in the literature and ease of re-implementation.\\
\textbf{Deep feature-based S-MAD:} Two different methods that include AlexNet and ResNet101 features are considered. These methods are derived from \cite{Venkatesh_2020_WACV} where the pre-trained AlexNet and ResNet101 features are extracted from the last layer, respectively, and final classification is performed using a linear SVM.  \\
\textbf{Hand-crafted feature-based S-MAD:} We employ hybrid feature based \cite{RagISBA2019} and ensemble feature-based \cite{EnsembleFeatures_2020} under hand-crafted feature category for our analysis.  The hybrid feature based S-MAD is based on the use of scale-space texture features classified using SRKDA classifier \cite{RagISBA2019}. The ensemble-based S-MAD is based on multiple features such as HoG, BSIF and LBP whose individual scores are obtained using P-CRC classifiers combined using sum rule to make the final decision. \\
\textbf{Morphing noise-based S-MAD:} We further employ two denoising based MAD approaches, namely, colour denoising \cite{DeepResidualNoise_IPTA_2019} and Residual Noise \cite{Venkatesh_2020_WACV}. The colour denoising method extracts the residual morphing noises in the HSV color space which is further classified using scale-space features and SRKDA classifiers. The residual noise-based method \cite{Venkatesh_2020_WACV}  is based on the Context Aggregated Network (CAN) that is used to perform the image denoising to extract the residual noise from the morphed image, which is further classified using P-CRC to detect the morphing images. 

\section{Experiments and Results}
\label{sec:exp}

This section presents the comprehensive evaluation of the six different MAD methods in three different evaluation protocols on four other ethnic groups. The performance of S-MAD is benchmarked using  ISO/IEC 30107-3  metrics such as  Attack Presentation Classification Error Rate (APCER (\%)), Bona fide Presentation Classification Error Rate (BPCER(\%)) and Detection-Equal Error Rate (D-EER(\%)). Further, we will also benchmark the MAD performance using Fairness Discrepancy Rate (FDR(\%)) to quantitatively benchmark the  S-MAD technique's fairness.

\begin{figure*}[htbp]
\begin{subfigure}{.245\textwidth}
  \centering
  \includegraphics[width=1\linewidth]{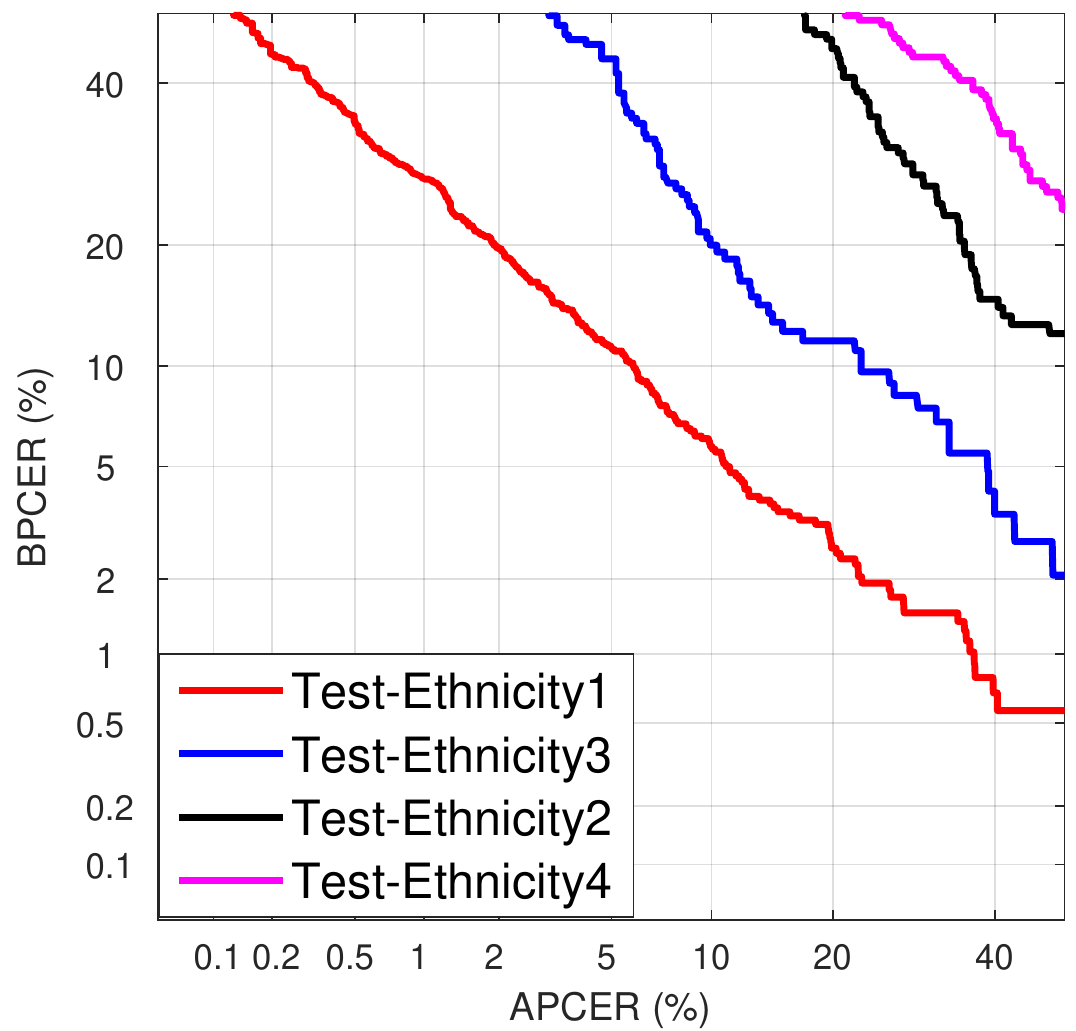}  
  \caption{Train: Ethnicity-1}
  \label{fig:sub-first}
\end{subfigure}
\begin{subfigure}{.245\textwidth}
  \centering
  \includegraphics[width=1\linewidth]{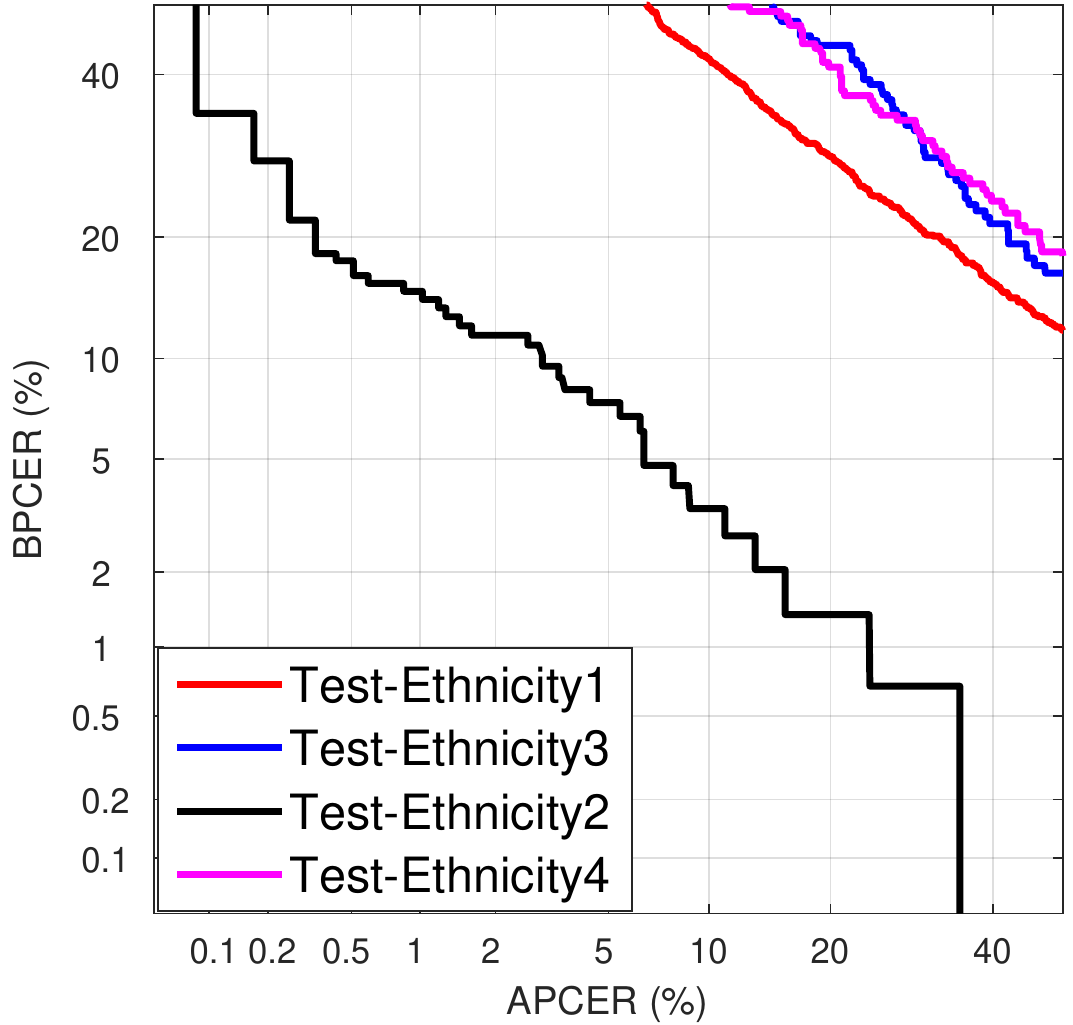}  
  \caption{Train: Ethnicity-2}
  \label{fig:sub-second}
\end{subfigure}
\begin{subfigure}{.245\textwidth}
  \centering
  \includegraphics[width=1\linewidth]{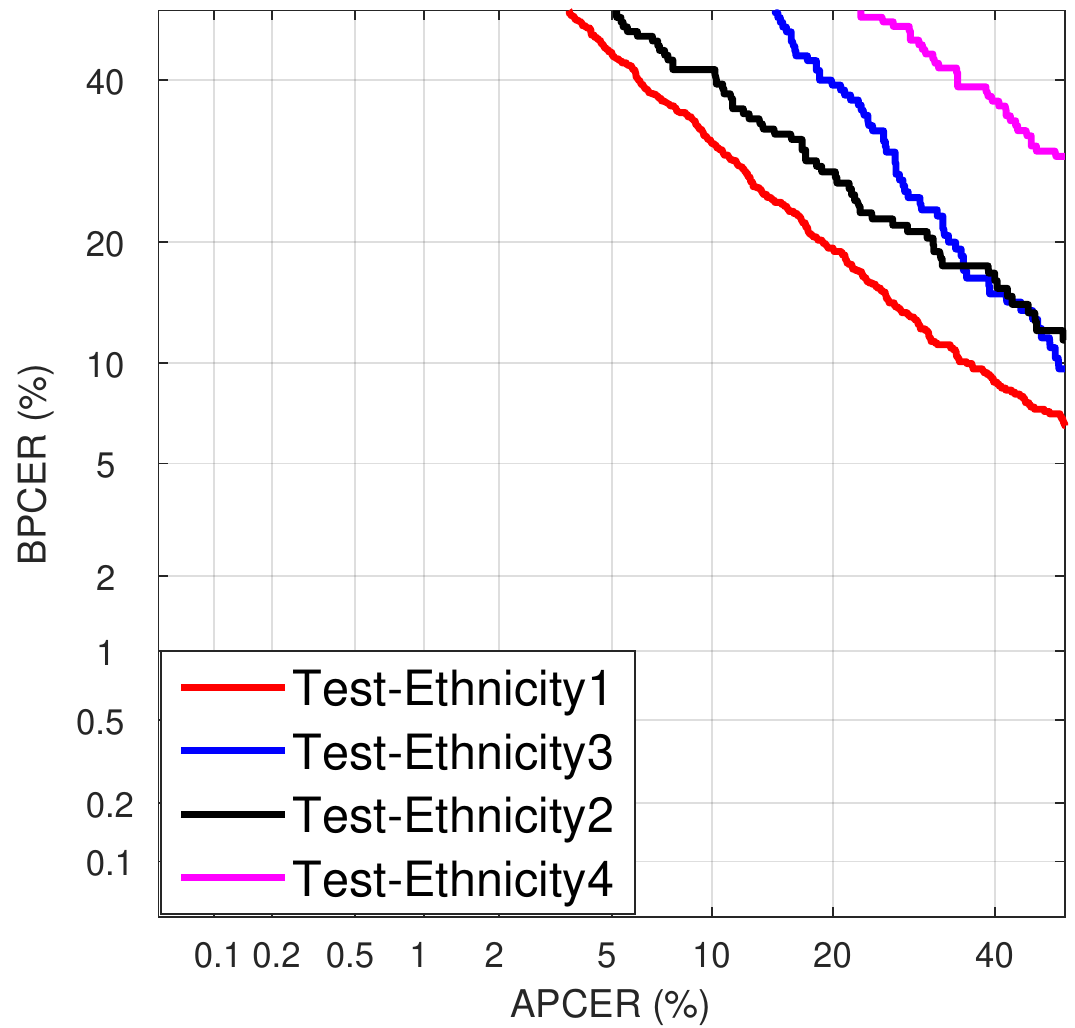}  
  \caption{Train: Ethnicity-3}
  \label{fig:sub-second}
\end{subfigure}
\begin{subfigure}{.245\textwidth}
  \centering
  \includegraphics[width=1\linewidth]{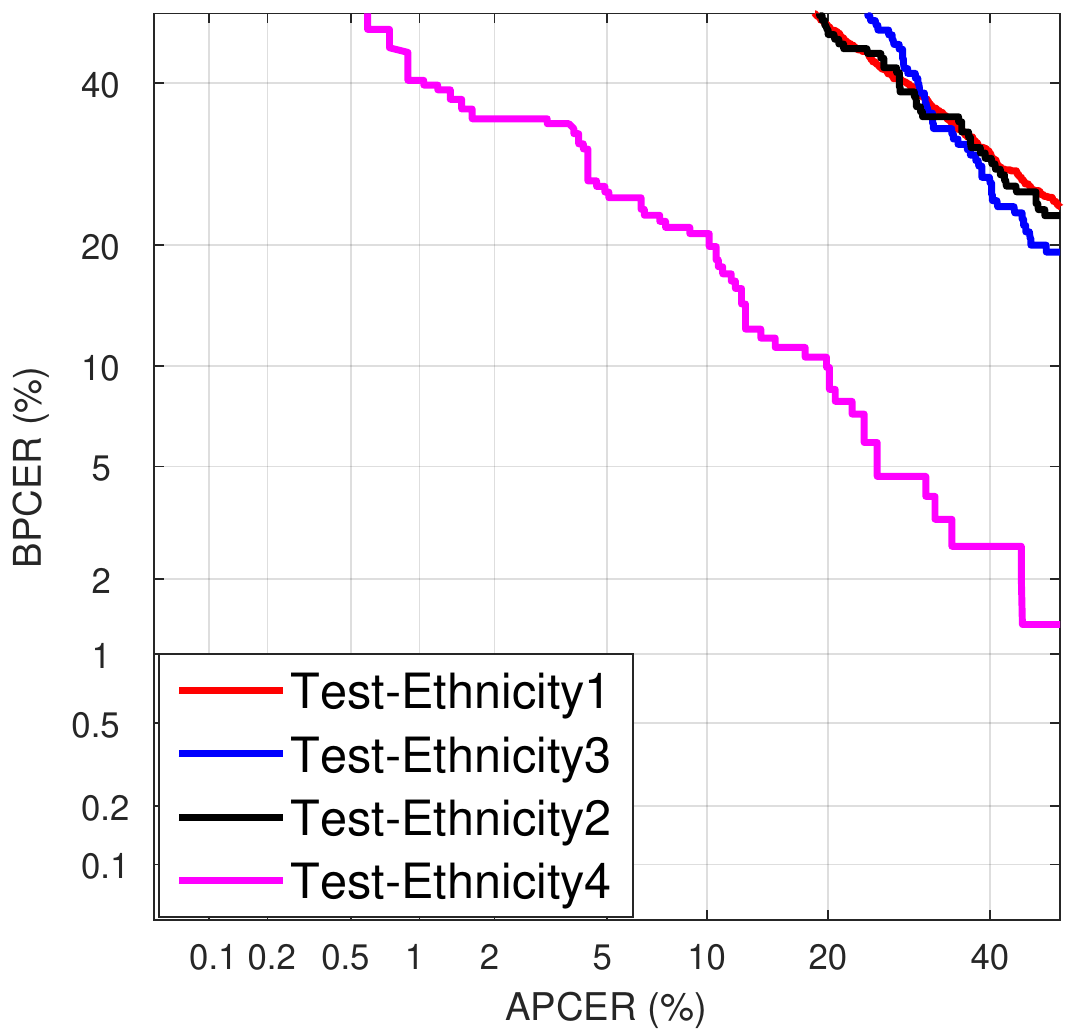}  
  \caption{Train: Ethnicity-4}
  \label{fig:sub-second}
\end{subfigure}
\caption{DET plots indicating the performance of Hybrid Features  \cite{RagISBA2019}}
\label{fig:Plot2}
\end{figure*}

\subsection{Experimental  Evaluation Protocol}
In this work, we have designed three different evaluation protocols: (1) \textbf{Intra-ethnicity evaluation:} This protocol provides insights into the performance achieved by individual S-MAD techniques on different ethnic groups. Here, the S-MAD techniques are trained and tested on the same ethnic groups. (2)  \textbf{Inter-ethnicity evaluation:} This protocol is designed to evaluate the S-MAD techniques in cross ethnicity scenarios. Here, the S-MAD technique is trained solely on the one type of ethnicity and tested on the remaining ethnic groups. (3) \textbf{Train-all-test-all evaluation:} This protocol evaluates the causality of the S-MAD techniques by training and testing the MAD techniques with all ethnic groups.

\subsection{Intra and Inter-Ethnicity Evaluation Results}
Table \ref{tab:TableEh1} - \ref{tab:TableEh4} shows the quantitative results of six different MAD techniques. In general, all six S-MAD techniques have indicated better detection accuracy, especially in the intra-ethnicity experiments. Among the six different S-MAD techniques, the residual noise-based method \cite{Venkatesh_2020_WACV} show the best performance on two different ethnic groups (Ethnicity-1 and Ethnicity-4). The ensemble features \cite{EnsembleFeatures_2020} indicate the best performance on Ethnicity-2, and AlexNet \cite{Venkatesh_2020_WACV} indicate the best performance with Ethnicity-3. It should however be noted that the performance of the individual S-MAD techniques varies across different ethnic groups.

\begin{table*}[htbp]
	\centering
	\caption{Quantitative  results  of FDR across different combination of ethnic groups in training-testing data.}
	\resizebox{0.62\textwidth}{!}{
		\begin{tabular}{lcccccc}
			\hline
			& \multicolumn{2}{c}{Eth-1 \& Eth-2} & \multicolumn{2}{c}{Eth-1 \& Eth-3} & \multicolumn{2}{c}{Eth-1 \& Eth-4} \bigstrut\\
			\hline
			& \multicolumn{2}{c}{FDR(\%) @ APCER} & \multicolumn{2}{c}{FDR(\%) @ APCER} & \multicolumn{2}{c}{FDR(\%) @ APCER} \bigstrut\\
			MAD Algorithm & 10\%  & 5\%  & 10\%  & 5\%  & 10\%  & 5\% \bigstrut\\
			\hline
			\hline
			Hybrid Features  \cite{RagISBA2019} & \textbf{98.81} & \textbf{98.12} & 41.65 & 33.28 & \textbf{84.59} & \textbf{85.41} \bigstrut\\
			Ensemble Features \cite{EnsembleFeatures_2020} & 95.35 & 93.58 & 82.42 & 74.28 & 87.47 & 83.21 \bigstrut\\
			Colour denoise \cite{DeepResidualNoise_IPTA_2019} & 92.73 & 88.48 & 69.09 & 63.69 & \textbf{97.48} & \textbf{96.35}  \bigstrut\\
			Residual Noise \cite{Venkatesh_2020_WACV} & 91.01 & 87.92 & \textbf{81.69} & \textbf{77.45} & \textbf{90.58} & \textbf{89.97} \bigstrut\\
			AlexNet \cite{Venkatesh_2020_WACV} & 91.38 & 95.36 & 84.29 & 74.01 & 86.47 & 80.74 \bigstrut\\
			ResNet101 \cite{Venkatesh_2020_WACV} & 93.48 & 91.94 & 73.67 & 69.01 & 72.39 & 54.87 \bigstrut\\
			\hline
			& \multicolumn{2}{c}{Eth-2 \& Eth-3} & \multicolumn{2}{c}{Eth-2 \& Eth-4} & \multicolumn{2}{c}{Eth-3 \& Eth-4} \bigstrut\\
			\hline
			Hybrid Features  \cite{RagISBA2019} & 39.29 & 29.15 & \textbf{82.21} & \textbf{81.66} & 57.77 & 47.89 \bigstrut\\
			Ensemble Features \cite{EnsembleFeatures_2020} & 73.1 & 61.46 & 78.17 & 70.36 & \textbf{94.93} & \textbf{91.1} \bigstrut\\
			Colour denoise \cite{DeepResidualNoise_IPTA_2019} & 54.56 & 40.63 & 82.94 & 77.61 & 71.62 & 63.02 \bigstrut\\
			Residual Noise \cite{Venkatesh_2020_WACV} & \textbf{99.68} & \textbf{98.39} & 91.44 & 47.47 & 91.12 & 87.48 \bigstrut\\
			AlexNet \cite{Venkatesh_2020_WACV} & 90.19 & 83.78 & \textbf{92.36} & \textbf{90.03} & 97.83 & 93.27 \bigstrut\\
			ResNet101 \cite{Venkatesh_2020_WACV} & \textbf{86.7} & \textbf{85.12} & 85.42 & 70.98 & 98.78 & 85.86 \bigstrut\\
			\hline    
		\end{tabular}%
	}
	\label{tab:FDR}%
\end{table*}%

The detection performance of S-MAD techniques are degraded in the Inter-ethnicity evaluation. The highest degradation is noted for deep learning-based S-MAD techniques. However, the residual noise-based S-MAD method \cite{Venkatesh_2020_WACV} indicates less degradation as compared to the other S-MAD techniques. Thus, this degradation can be attributed to variations in ethnicity and the data quality due to the different capturing conditions.  Further, it is interesting to note that, when trained on Ethnicity-3, the performance of the MAD techniques with inter and intra evaluation protocol did not show a significant difference. Thus, S-MAD methods indicate the influence of the ethnicity distribution in trained data. Fig~\ref{fig:Plot1} and Fig~\ref{fig:Plot2} shows the DET curves for residual noise-based method \cite{Venkatesh_2020_WACV} and Hybrid features \cite{RagISBA2019} respectively. For simplicity, we have included the DET plots corresponding to these two MAD techniques. These DET plots clearly show the improved detection performance of the S-MAD techniques with intra evaluation protocol as compared to inter evaluation protocol.

Table \ref{tab:TableEh5} indicates the detection performance of six different S-MAD methods that are trained with four different types of ethnicity. The obtained results suggest that the detection performed is greatly improved when compared to the cross-ethnicity scenarios. However, the performance of the S-MAD algorithms is influenced by the ethnicity type of the testing data.  The deep feature-based S-MAD algorithms indicate degraded results indicating the lack of generalization. 

\subsection{Fairness Measure}
In this section, we present the fairness of the MAD methods on all four different ethnic groups. To this extent, we employ the Fairness Discrepancy Rate (FDR) \cite{de2020fairness} to compute the quantitative value of the  fairness. The FDR metric is based on False Match Rate (FMR) and False Non-Match Rate (FNMR) and initially proposed to benchmark face recognition systems. In this work, we adapt the same metric in terms of APCER and BPCER to compute the FDR as follows: 

\begin{equation}
FDR = 1 - (\alpha A(\tau) + (1 - \alpha)B(\tau))
\end{equation}
Where, $\tau$ represents the operating threshold of the MAD technique, $\alpha$ represents the hyper-parameter that defines the weight of $A(\tau)$ and $B(\tau)$. The $A(\tau) = \max(\vert APCER(\tau)^{d_{i}} -  APCER(\tau)^{d_{j} } \vert)$, $B(\tau) = \max(\vert BPCER(\tau)^{d_{i}} -  BPCER(\tau)^{d_{j} } \vert)$, $D = {d_{i}, d_{j}}$ indicates the demographic groups. The algorithm is said to be fair, if the performance at various  operating threshold remains constant.  

We have benchmarked the fairness of six different MAD techniques trained and tested on the same ethnic group. Therefore, to present the comprehensive comparison across four different ethnic groups, we computed the fairness measure across two ethnic groups at a time. Table \ref{tab:FDR} indicates the quantitative value of the FDR corresponding to six different S-MAD techniques. Based on the obtained results, it can be noted that: (1) All six different S-MAD techniques fail to show a stable detection performance across different operating threshold set at APCER = 5\% and 10\% with different ethnic groups. (2) The detection performance of the individual S-MAD algorithms varies with the ethnicity type used to train the S-MAD techniques (3) State-of-the-art S-MAD techniques fail to demonstrate fairness across different ethnic groups.

\section{Discussion}
\label{sec:discussion}
Based on the analysis conducted, obtained results and the observations made above, the research questions formulated in Section \ref{sec:intro} are answered below.
\begin{itemize}[leftmargin=*,noitemsep, topsep=0pt,parsep=0pt,partopsep=0pt]
	\item {{\textbf{Q1}}. Does the ethnicity influence the detection performance of S-MAD algorithms?}
	\begin{itemize}[leftmargin=*,noitemsep, topsep=0pt,parsep=0pt,partopsep=0pt]
		\item As noted by the obtained experimental results reported in Table \ref{tab:TableEh1} - \ref{tab:FDR}, existing S-MAD methods detection performance are influenced by the ethnicity of subjects used for both training and testing). 
	\end{itemize}
	\item {\textbf{Q2}. For which ethnic groups is the detection most challenging? } 
	\begin{itemize}[leftmargin=*,noitemsep, topsep=0pt,parsep=0pt,partopsep=0pt]
		\item As noted from the experimental results (see Table Table \ref{tab:TableEh1} - \ref{tab:FDR}), it is difficult to conclude which ethnic group is showing a degraded detection performance. In general, ethnic groups together with the data capture environment will influence the fairness of the S-MAD techniques across different ethnic groups. However, the individual S-MAD algorithms indicates the variation in performance across different ethnic groups. 
	\end{itemize}
	\item {\textbf{Q3}.What category of S-MAD algorithms (deep learning/hand-crafted features/morphing noise based) degrade in performance as a result of ethnic bias?}
	\begin{itemize}[leftmargin=*,noitemsep, topsep=0pt,parsep=0pt,partopsep=0pt]
		\item Based on the experimental results (see Table \ref{tab:TableEh1} - \ref{tab:FDR}), deep-features based S-MAD techniques have indicated the degraded detection  performance across different ethnic groups.  The best detection performance is noted for the residual noise-based method \cite{Venkatesh_2020_WACV}. 
	\end{itemize}           
\end{itemize}
\subsection{Limitations of current work}
\label{ssec:limitations}
While the new dataset of mixed ethnicity distribution answers some of the fundamental questions, one should also consider the role of varying data capture conditions. If a larger dataset of different ethnicities can be captured under the same conditions with the same set of capture devices, one can eliminate any other factors and purely study the algorithmic bias towards ethnicity. This aspect should be considered in future works. Secondly, a detailed analysis of learned features with same classifier should also be studied for understanding the invariance of different features in MAD with respect to ethnicity in the future works.

\section{Conclusion}
\label{sec:conc}
Morphing attacks can successfully attack FRS and in order to mitigate this risk, a number of MAD algorithms are proposed. Even though the influence of ethnic groups resulting in a FRS bias is well studied in the literature, a similar study is not yet presented for face morphing attack detection. This work presents an extensive analysis to benchmark the quantitative detection performance of six different S-MAD techniques, specifically with respect to variety of ethnicity in the training data. To this extent, a mixed ethnicity morphed dataset was constructed using four different ethnic groups (including a newly collected face dataset referred to as Ethnicity-4).  An extensive evaluation is carried out using three different evaluation protocols to benchmark the detection performance bias due to ethnicity. The fairness of the S-MAD techniques is quantified using Fairness Discrepancy Rate (FDR) metric. Based on the obtained results, the state-of-the-art S-MAD techniques evaluated in this work lack algorithmic fairness towards different ethnic groups. The need for not only robust, but better algorithms are underlined to mitigate any negative social, legal and ethical implications should such MAD algorithms be deployed.

\subsection{Acknowledgment}
\label{ssec:limitations}
Authors like to thank Sushma Venkatesh for her constructive feedback.  
\balance
{\small
		\bibliographystyle{ieee}
	\bibliography{Paper_V1}

\begin{thebibliography}{10}\itemsep=-1pt

\bibitem{CASIAFAceDB}
Casia vis-nir-3d dataset.
\newblock \url{http://www.cbsr.ia.ac.cn/english/NIR-VIS-2.0-Database.html}.
\newblock Accessed: 20, April 2020.

\bibitem{aghdaie2021attention}
P.~Aghdaie, B.~Chaudhary, S.~Soleymani, J.~Dawson, and N.~M. Nasrabadi.
\newblock Attention aware wavelet-based detection of morphed face images.
\newblock In {\em 2021 IEEE International Joint Conference on Biometrics
  (IJCB)}, pages 1--8. IEEE, 2021.

\bibitem{banerjee2021conditional}
S.~Banerjee and A.~Ross.
\newblock Conditional identity disentanglement for differential face morph
  detection.
\newblock In {\em 2021 IEEE International Joint Conference on Biometrics
  (IJCB)}, pages 1--8. IEEE, 2021.

\bibitem{MORPHII_DB}
G.~Bingham, K.~Kempfert, B.~Yip, J.~Fabish, M.~Ferguson, C.~Nansalo, K.~Park,
  R.~Towner, T.~Kling, Y.~Wang, et~al.
\newblock Preliminary studies on a large face database morph-ii.

\bibitem{damer2021regenmorph}
N.~Damer, K.~Raja, M.~S{\"u}{\ss}milch, S.~Venkatesh, F.~Boutros, M.~Fang,
  F.~Kirchbuchner, R.~Ramachandra, and A.~Kuijper.
\newblock Regenmorph: Visibly realistic gan generated face morphing attacks by
  attack re-generation.
\newblock {\em arXiv preprint arXiv:2108.09130}, 2021.

\bibitem{RightFaces_Naser_2019}
N.~{Damer}, A.~M. {Saladié}, S.~{Zienert}, Y.~{Wainakh}, P.~{Terhörst},
  F.~{Kirchbuchner}, and A.~{Kuijper}.
\newblock To detect or not to detect: The right faces to morph.
\newblock In {\em 2019 International Conference on Biometrics (ICB)}, pages
  1--8, 2019.

\bibitem{damer2021pw}
N.~Damer, N.~Spiller, M.~Fang, F.~Boutros, F.~Kirchbuchner, and A.~Kuijper.
\newblock Pw-mad: Pixel-wise supervision for generalized face morphing attack
  detection.
\newblock {\em arXiv preprint arXiv:2108.10291}, 2021.

\bibitem{de2020fairness}
T.~de~Freitas~Pereira and S.~Marcel.
\newblock Fairness in biometrics: a figure of merit to assess biometric
  verification systems.
\newblock {\em arXiv preprint arXiv:2011.02395}, 2020.

\bibitem{MargicPass}
M.~Ferrara, A.~Franco, and D.~Maltoni.
\newblock The magic passport.
\newblock In {\em Biometrics (IJCB), 2014 IEEE International Joint Conference
  on}, pages 1--7, 2014.

\bibitem{Ferrara2016}
M.~Ferrara, A.~Franco, and D.~Maltoni.
\newblock {\em Face Recognition Across the Imaging Spectrum}, chapter On the
  Effects of Image Alterations on Face Recognition Accuracy, pages 195--222.
\newblock Springer International Publishing, 2016.

\bibitem{UBO_Morphing_Tool}
M.~Ferrara, A.~Franco, and D.~Maltoni.
\newblock Decoupling texture blending and shape warping in face morphing.
\newblock In {\em 2019 International Conference of the Biometrics Special
  Interest Group (BIOSIG)}, pages 1--5. IEEE, 2019.

\bibitem{IWBF2017_StirTrace}
M.~Hildebrandt, T.~Neubert, A.~Makrushin, and J.~Dittmann.
\newblock Benchmarking face morphing forgery detection: Application of
  stirtrace for impact simulation of different processing steps.
\newblock In {\em International Workshop on Biometrics and Forensics (IWBF
  2017)}, pages 1--6, 2017.

\bibitem{howard2020quantifying}
J.~J. Howard, Y.~B. Sirotin, J.~L. Tipton, and A.~R. Vemury.
\newblock Quantifying the extent to which race and gender features determine
  identity in commercial face recognition algorithms.
\newblock {\em arXiv preprint arXiv:2010.07979}, 2020.

\bibitem{MAD_Human2019}
R.~S.~S. Kramer, M.~O. Mireku, T.~Flack, and K.~L. Ritchie.
\newblock Face morphing attacks: Investigating detection with humans and
  computers.
\newblock {\em Cognitive Research: Principles and Implications}, 4, 2019.

\bibitem{MakrushinHuman}
A.~Makrushin., T.~Neubert., and J.~Dittmann.
\newblock Humans vs. algorithms: Assessment of security risks posed by facial
  morphing to identity verification at border control.
\newblock In {\em Proceedings of the 14th International Joint Conference on
  Computer Vision, Imaging and Computer Graphics Theory and Applications -
  Volume 4: VISAPP,}, pages 513--520. INSTICC, SciTePress, 2019.

\bibitem{makrushin2018overview}
A.~Makrushin and A.~Wolf.
\newblock An overview of recent advances in assessing and mitigating the face
  morphing attack.
\newblock In {\em 2018 26th European Signal Processing Conference (EUSIPCO)},
  pages 1017--1021. IEEE, 2018.

\bibitem{Ferrara_PrintScan2019}
F.~Matteo, F.~Annalisa, and M.~Davide.
\newblock Face morphing detection in the presence of printing/scanning and
  heterogeneous image sources.
\newblock In {\em ArXiv}.

\bibitem{mehrabi2021survey}
N.~Mehrabi, F.~Morstatter, N.~Saxena, K.~Lerman, and A.~Galstyan.
\newblock A survey on bias and fairness in machine learning.
\newblock {\em ACM Computing Surveys (CSUR)}, 54(6):1--35, 2021.

\bibitem{neubert2017Imagedegradation}
T.~Neubert.
\newblock Face morphing detection: An approach based on image degradation
  analysis.
\newblock In {\em International Workshop on Digital Watermarking}, pages
  93--106.

\bibitem{NISTReport2020}
M.~Ngan, P.~Grother, K.~Hanaoka, and J.~Kuo.
\newblock Face recognition vendor test (frvt) part 4: Morph-performance of
  automated face morph detection.
\newblock {\em National Institute of Technology (NIST), Tech. Rep. NISTIR},
  8292, 2021.

\bibitem{raghavendra2017face}
R.~Raghavendra, K.~Raja, S.~Venkatesh, and C.~Busch.
\newblock Face morphing versus face averaging: Vulnerability and detection.
\newblock In {\em IEEE International Joint Conference on Biometrics (IJCB)},
  pages 555--563, 2017.

\bibitem{raghavendra2017transferable}
R.~Raghavendra, K.~Raja, S.~Venkatesh, and C.~Busch.
\newblock Transferable deep-cnn features for detecting digital and
  print-scanned morphed face images.
\newblock In {\em Proc. IEEE Conf. Computer Vision Pattern Recognition
  Workshops (CVPRW)}, pages 1822--1830, 2017.

\bibitem{Raghavendra2016}
R.~Raghavendra, K.~B. Raja, and C.~Busch.
\newblock {Detecting Morphed Face Images}.
\newblock In {\em 8th IEEE International Conference on Biometrics: Theory,
  Applications, and Systems (BTAS)}, pages 1--8, 2016.

\bibitem{RagCVIP2018}
R.~Raghavendra, S.~Venkatesh, K.~Raja, and C.~Busch.
\newblock Detecting face morphing attacks with collaborative representation of
  steerable features.
\newblock In {\em IAPR International Conference on Computer Vision \& Image
  Processing (CVIP-2018)}, pages 1--7, 2018.

\bibitem{RagISBA2019}
R.~Raghavendra, S.~Venkatesh, K.~Raja, and C.~Busch.
\newblock Towards making morphing attack detection robust using hybrid
  scale-space colour texture features.
\newblock In {\em IEEE International Conference on Identity, Security and
  Behaviour Analysis (ISBA 2019)}, pages 1--7, 2019.

\bibitem{Raja2020MorphingAD}
K.~Raja, M.~Ferrara, A.~Franco, L.~J. Spreeuwers, I.~Batskos, F.~de~Wit,
  M.~Gomez-Barrero, U.~Scherhag, D.~Fischer, S.~Venkatesh, J.~M. Singh, G.~Li,
  L.~Bergeron, S.~Isadskiy, R.~Raghavendra, C.~Rathgeb, D.~Frings, U.~Seidel,
  F.~Knopjes, R.~N.~J. Veldhuis, D.~Maltoni, and C.~Busch.
\newblock Morphing attack detection - database, evaluation platform and
  benchmarking.
\newblock {\em IEEE Transactions on Information Forensics and Security (TIFS)},
  3, 2020.

\bibitem{Morph_PLUSONE}
D.~Robertson, R.~S. Kramer, and A.~M. Burton.
\newblock Fraudulent id using face morphs: Experiments on human and automatic
  recognition.
\newblock {\em PloS ONE}, 12(3):1--12, 2017.

\bibitem{RobertSon2018}
D.~J. Robertson, A.~Mungall, D.~G. Watson, K.~A. Wade, S.~J. Nightingale, and
  S.~Butler.
\newblock Detecting morphed passport photos: a training and individual
  differences approach.
\newblock {\em Cognitive Research: Principles and Implications}, 3(27):1--12,
  2018.

\bibitem{robinson2020face}
J.~P. Robinson, G.~Livitz, Y.~Henon, C.~Qin, Y.~Fu, and S.~Timoner.
\newblock Face recognition: too bias, or not too bias?
\newblock In {\em Proceedings of the IEEE/CVF Conference on Computer Vision and
  Pattern Recognition Workshops}, pages 0--1, 2020.

\bibitem{Scher2017}
U.~Scherhag, R.~Raghavendra, K.~Raja, M.~Gomez-Barrero, C.~Rathgeb, and
  C.~Busch.
\newblock On the vulnerability of face recognition systems towards morphed face
  attack.
\newblock In {\em International Workshop on Biometrics and Forensics (IWBF
  2017)}, pages 1--6, 2017.

\bibitem{scherhag2020deep}
U.~{Scherhag}, C.~{Rathgeb}, J.~{Merkle}, and C.~{Busch}.
\newblock Deep face representations for differential morphing attack detection.
\newblock {\em IEEE Transactions on Information Forensics and Security},
  15:3625--3639, 2020.

\bibitem{seibold2018reflection}
C.~Seibold, A.~Hilsmann, and P.~Eisert.
\newblock Reflection analysis for face morphing attack detection.
\newblock {\em arXiv preprint arXiv:1807.02030}, 2018.

\bibitem{Jag_ABC_gate_2019}
J.~M. {Singh}, R.~{Raghavendra}, K.~B. {Raja}, and C.~{Busch}.
\newblock Robust morph-detection at automated border control gate using deep
  decomposed 3d shape diffuse reflectance.
\newblock In {\em 2019 15th International Conference on Signal-Image Technology
  Internet-Based Systems (SITIS)}, pages 106--112, 2019.

\bibitem{soleymani2020differential}
S.~Soleymani, B.~Chaudhary, A.~Dabouei, J.~Dawson, and N.~M. Nasrabadi.
\newblock Differential morphed face detection using deep siamese networks.
\newblock {\em arXiv preprint arXiv:2012.01541}, 2020.

\bibitem{soleymani2021mutual}
S.~Soleymani, A.~Dabouei, F.~Taherkhani, J.~Dawson, and N.~M. Nasrabadi.
\newblock Mutual information maximization on disentangled representations for
  differential morph detection.
\newblock In {\em Proceedings of the IEEE/CVF Winter Conference on Applications
  of Computer Vision}, pages 1731--1741, 2021.

\bibitem{terhorst2021comprehensive}
P.~Terh{\"o}rst, J.~N. Kolf, M.~Huber, F.~Kirchbuchner, N.~Damer, A.~Morales,
  J.~Fierrez, and A.~Kuijper.
\newblock A comprehensive study on face recognition biases beyond demographics.
\newblock {\em arXiv preprint arXiv:2103.01592}, 2021.

\bibitem{EnsembleFeatures_2020}
S.~{Venkatesh}, R.~{Raghavendra}, K.~{Raja}, and C.~{Busch}.
\newblock Single image face morphing attack detection using ensemble of
  features.
\newblock In {\em 23rd International Conference on Information Fusion}, pages
  1--5, 2020.

\bibitem{Venkatesh_2020_WACV}
S.~Venkatesh, R.~Raghavendra, K.~Raja, L.~Spreeuwers, R.~Veldhuis, and
  C.~Busch.
\newblock Detecting morphed face attacks using residual noise from deep
  multi-scale context aggregation network.
\newblock In {\em The IEEE Winter Conference on Applications of Computer Vision
  (WACV)}, pages 1--8, March 2020.

\bibitem{DeepResidualNoise_IPTA_2019}
S.~Venkatesh, R.~Raghavendra, K.~Raja, L.~Spreeuwers, R.~Veldhuis, and
  C.~Busch.
\newblock Morphed face detection based on deep color residual noise.
\newblock In {\em International Conference on Image Processing, Theory, Tools
  and Applications (IPTA)}, pages 1--8, November 2019.

\bibitem{Venkatesh_2020_IJCB}
S.~Venkatesh, K.~Raja, R.~Raghavendra, and C.~Busch.
\newblock On the influence of ageing on face morph attacks: Vulnerability and
  detection.
\newblock In {\em International Joint Conference on Biometrics (IJCB)}, pages
  1--8, September 2020.

\bibitem{Sushma_MAD_Survey}
S.~{Venkatesh}, R.~{Ramachandra}, K.~{Raja}, and C.~{Busch}.
\newblock Face morphing attack generation \& detection: A comprehensive survey.
\newblock {\em IEEE Transactions on Technology and Society}, pages 1--23, 2021.

\bibitem{MorphStyleGAN2020}
S.~Venkatesh, H.~Zhang, R.~Raghavendra, K.~Raja, N.~Damer, and C.~Busch.
\newblock Can gan generated morphs threaten face recognition systems equally as
  landmark based morphs? - vulnerability and detection.
\newblock In {\em 2020 International Workshop on Biometrics and Forensics
  (IWBF)}, pages 1--6. IEEE, 2020.

\bibitem{MIPGAN}
H.~Zhang, S.~Venkatesh, R.~Ramachandra, K.~Raja, N.~Damer, and C.~Busch.
\newblock {MIPGAN}--generating robust and high quality morph attacks using
  identity prior driven {GAN}.
\newblock {\em IEEE Transactions on Biometrics, Behavior, and Identity Science
  (TBIOM)}, 2, 2021.

\end{thebibliography}
}

\end{document}